\pdfoutput=1
\documentclass[11pt]{article}

\providecommand{\tightlist}{%
  \setlength{\itemsep}{0pt}\setlength{\parskip}{0pt}}
  
\usepackage{color}
\usepackage{graphicx}
\usepackage{color,soul} 

\usepackage{amsmath}
\usepackage{ACL2023}

\usepackage{times}
\usepackage{latexsym}
\usepackage{booktabs}

\usepackage[T1]{fontenc}

\usepackage[utf8]{inputenc}

\usepackage{microtype}

\usepackage{inconsolata}

\title{CLaC at SemEval-2025 Task 6: A Multi-Architecture Approach for Corporate Environmental Promise Verification}

\author{Nawar Turk$^*$
\And  Eeham Khan$^*$
\And Leila Kosseim
\AND\vspace{-2.25em}\\
        Computational Linguistics at Concordia (CLaC) Laboratory \\
        Department of Computer Science and Software Engineering \\
        Concordia University, Montr\'eal, Qu\'ebec, Canada \\
    \texttt{\{nawar.turk,eeham.khan,leila.kosseim\}@concordia.ca}
        }

\begin{document}
\maketitle
\renewcommand{\thefootnote}{\fnsymbol{footnote}}
\footnotetext[1]{Equal contribution.}
\renewcommand{\thefootnote}{\arabic{footnote}}
\begin{abstract}
This paper presents our approach to the  SemEval-2025 Task~6 (PromiseEval), which focuses on verifying promises in corporate ESG (Environmental, Social, and Governance) reports. We explore three model architectures to address the four subtasks of promise identification, supporting evidence assessment, clarity evaluation, and verification timing. Our first model utilizes ESG-BERT with task-specific classifier heads, while our second model enhances this architecture with linguistic features tailored for each subtask. Our third approach implements a combined subtask model with attention-based sequence pooling, transformer representations augmented with document metadata, and multi-objective learning. Experiments on the English portion of the ML-Promise dataset demonstrate progressive improvement across our models, with our combined subtask approach achieving a leaderboard score of 0.5268, outperforming the provided baseline of 0.5227. Our work highlights the effectiveness of linguistic feature extraction, attention pooling, and multi-objective learning in promise verification tasks, despite challenges posed by class imbalance and limited training data.
\end{abstract}

\section{Introduction}
The PromiseEval task at SemEval-2025~\cite{chen-etal-2025-semeval} addresses the critical challenge of verifying promises in ESG (Environmental, Social, and Governance) reports published by corporations across multiple languages and industries. Corporate promises significantly influence stakeholder trust and organizational reputation, yet their complexity and volume make verification difficult. This task breaks down promise verification into four essential subtasks: 
\begin{enumerate}\tightlist
    \item \textbf{Promise Identification:} Determining if a statement contains a promise or not.
    \item \textbf{Supporting Evidence Assessment:} Verifying if the promise has concrete evidence or not.
    \item \textbf{Clarity of the Promise-Evidence Pair:}  Classifying the promise evidence as Clear, Not Clear, or Misleading.
    \item \textbf{Timing for Verification:} Categorizing when promises should be verified within~2 years,~2-5 years, beyond 5 years, or other.
\end{enumerate}

All of the code used in the implementation of the models described in this paper is made available on \texttt{GitHub}\footnote{\label{github}\url{https://github.com/CLaC-Lab/SemEval-2025-Task6}}.

\section{Background}
The PromiseEval task at SemEval-2025 builds upon the ML-Promise dataset introduced by~\cite{seki2024mlpromise}, the first multilingual resource for promise verification in corporate ESG communications. For our experiments, we focused exclusively on the English portion containing ~400 training instances labelled for each of the four classification subtasks. 

\begin{figure}[htbp]
    \centering
    \includegraphics[width=1\columnwidth]{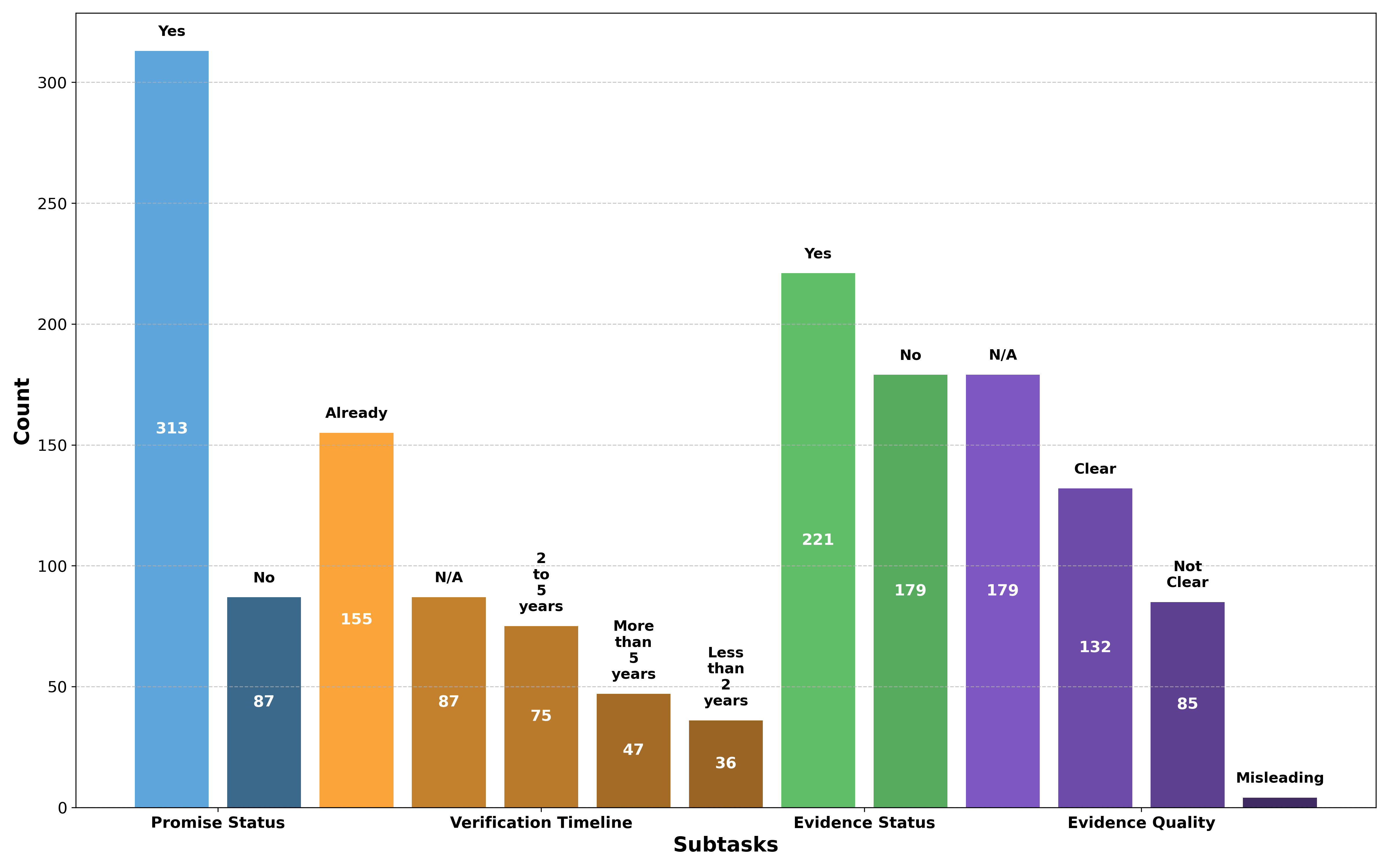} 
    \caption{Class distribution across four subtasks in the English portion of the SemEval-2025 Task 6 dataset.}
    \label{fig:datasetdistribution}
\end{figure}

As shown in Figure~\ref{fig:datasetdistribution}, the dataset exhibits class imbalance across all four subtasks, creating challenges for model development. Our work explores effective model architectures, with a particular focus on linguistic feature extraction and multi-objective learning with contextual enrichment. The combined subtask model addresses these challenges through attention-based sequence pooling, transformer representations augmented with document metadata, and a training methodology incorporating focal loss and test-time augmentation to improve performance on imbalanced classes while maintaining computational efficiency.

\section{Related Work}
Our work builds upon several interconnected areas in Natural Language Processing (NLP) and ESG text analysis. This section outlines prior work related to our approach and contextualizes our contributions.
\subsection{ESG Text Analysis}
The computational analysis of Environmental, Social, and Governance (ESG) disclosures has received growing attention in recent years.~\citet{armbrust-etal-2020-computational} developed a framework for analyzing corporate sustainability reports using NLP techniques, identifying key sustainability themes and measuring their prevalence across sectors. Similarly,~\citet{bingler2022cheap} examined the phenomenon of ``green-washing'' in corporate climate pledges, highlighting inconsistencies between commitments and actions.
The development of domain-specific language models has been particularly important for ESG text analysis.~\citet{mukherjee2021esgbert} introduced ESG-BERT, which we employ in our Base and Feature-Enhanced models (see Sections~\ref{sec:based} and~\ref{sec:enhanced}). 

\subsection{Multi-Task Learning in NLP}
Our Combined Subtask Model (see Section~\ref{sec:combined}) incorporates multi-task learning principles, which have shown their effectiveness in related NLP challenges.~\citet{liu2019multi} demonstrated that multi-task learning improves performance across various NLP tasks by enabling knowledge transfer between related classification objectives. Similarly,~\citet{chen2024multitasklearningnaturallanguage} provided a comprehensive overview of multi-task learning approaches in NLP, highlighting the benefits of shared representations for related tasks.

In the financial domain,~\citet{yang2020financial} employed multi-task learning for analyzing financial documents, jointly modelling document classification and named entity recognition tasks with a shared encoder. Their approach demonstrated performance improvements similar to our findings regarding joint promise and evidence detection.

\subsection{Attention Mechanisms and Feature Engineering}
The attention pooling mechanism implemented in our Combined Subtask Model (see Section~\ref{sec:combined}) draws inspiration from work by~\citet{yang-etal-2016-hierarchical} on hierarchical attention networks for document classification. Their approach demonstrated the effectiveness of attention mechanisms for focusing on relevant parts of documents, particularly for long texts like the corporate reports in our dataset. For linguistic feature engineering, our approach builds on work by~\citet{prabhakaran-etal-2016-predicting} who used linguistic markers to identify commitment language in political discourse. 

\subsection{Class Imbalance in Text Classification}
Class imbalance has been addressed by several researchers.~\citet{johnson2019survey} provided a comprehensive survey of techniques for handling imbalanced data in machine learning, several of which we incorporated in our approach. More specific to NLP,~\citet{henning-etal-2023-survey} explored techniques for addressing class imbalance in transformer-based text classification, demonstrating that appropriate loss functions and sampling strategies can significantly improve performance for minority classes.

Our test-time augmentation approach (see Section~\ref{sec:combined}) draws on work by~\citet{shanmugam2021betteraggregationtesttimeaugmentation}, who demonstrated that augmented inference can improve classification performance, particularly in challenging examples.

\section{System Overview}
\begin{figure*}[htbp]
    \centering
\includegraphics[width=.6\textwidth]{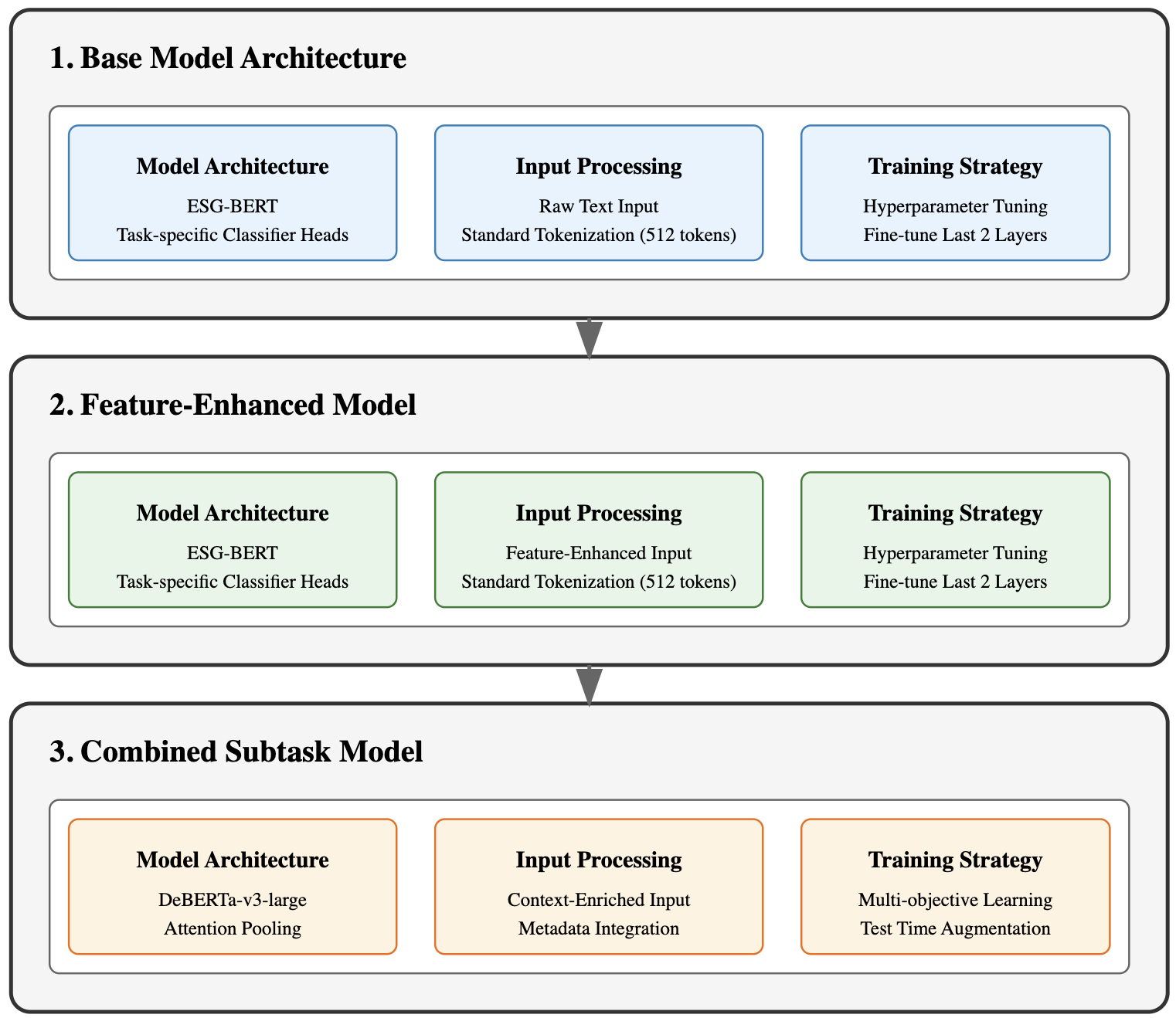} 
    \caption{System architecture of the three Promise Verification models.}
    \label{fig:systemarchitecture.png}
\end{figure*}

Our system explores three different model architectures, as illustrated in Figure~\ref{fig:systemarchitecture.png}.

\begin{description}\tightlist
    \item \textbf{Model~1: } a baseline architecture with ESG-BERT and task-specific classifier heads processing the given text inputs as is.
    \item \textbf{Model~2: } a feature-enhanced ESG-BERT model incorporating linguistic features tailored for each subtask.
    \item \textbf{Model~3: } a combined subtask model that integrates a multi-objective architecture for subtasks~1 and~2, and uses attention pooling with a shared transformer backbone for better feature extraction.
\end{description}

\subsection{Model~1: Base Model Architecture}
\label{sec:based} 
Our Base Model consisted of the pre-trained ESG-BERT~\cite {mukherjee2021esgbert} model with four subtask-specific classifier heads. We trained four distinct models, one for each subtask in the promise verification pipeline. ESG-BERT was selected for all subtasks to leverage its domain-specific knowledge of environmental, social, and governance terminology, which closely aligns with the content of corporate promise statements.
To optimize training efficiency while maintaining model performance with our limited dataset of~400 instances, we froze the ESG-BERT's model parameters and only fine-tuned the last~2 transformer layers along with the classification heads. This approach significantly reduced computational requirements and potentially helped prevent overfitting.

\subsection{Model~2: Feature-Enhanced Model}
\label{sec:enhanced}
For our second architecture, we enhanced the Base ESG-BERT model with explicit linguistic features tailored to each subtask. We made the assumption that prepending task-specific feature tags to the input text would improve model performance by signalling important linguistic patterns that might otherwise require many training examples to learn.

\paragraph{\textbf{Subtask~1 -- Promise Identification:}} We generated a list of promise-related terms (e.g., ``commit'', ``pledge'',  ``goal'') and prepended the presence of one of these (stemmed) terms to the input. In addition, we included the sentiment polarity of the input, based on the hypothesis that promises are typically expressed positively. For example, given the original text:

\noindent{\small{\texttt{We commit to achieving net-zero emissions across our entire supply chain by 2040}}}

Model~2 would transform it to:

\noindent{\small{\texttt{POSITIVE Sentiment. Contains Promise Word. We commit to achieving net-zero emissions across our entire supply chain by 2040}}}

\paragraph{\textbf{Subtask~2 -- Evidence Detection:}} We developed two sets of terms for concrete metrics (e.g., ``percentage'', ``dollars'') and supporting evidence (e.g.,  ``document''). We then prepended feature tags indicating the count of these terms and the presence of numbers and dates detected via named entity recognition (NER) models, as in:

\noindent{\small{\texttt{
Proof\_Count\_2. Has\_Numbers. Our carbon emissions decreased by 15\%, as stated in our sustainability report and confirmed through third-party audit}}}

\paragraph{\textbf{Subtask~3 -- Clarity Assessment:}}We crafted
two tailored lists of vague terms signalling evasive language, and of terms indicating clear language. We counted occurrences and prepended these counts to the texts. For example:

\noindent{\small{\texttt{Vague\_Terms\_2. Specific\_Terms\_0.  We might consider implementing sustainability initiatives}}}

\paragraph{\textbf{Subtask~4 -- Timing for Verification:}} We developed lists of time-related terms for four verification timeframes (e.g., 2-5 years, more than 5 years), extracted dates using NER, and prepended this information.

\subsection{Model~3: Combined Subtask Model}
\label{sec:combined}

For our third model, we implemented a multitask learning framework, specifically focusing on Subtasks~1 and~2. The core of our system is built on the DeBERTa-v3-large transformer model~\cite{he2021debertadecodingenhancedbertdisentangled}, with several architectural additions:

\paragraph{\textbf{Attention Pooling:}} Instead of relying on the standard [CLS] token representation, we implemented attention pooling to dynamically weight token representations across the sequence, allowing the model to better focus on relevant textual elements:
\begin{equation}
\alpha_i = \text{softmax}(W_{\text{attn}}h_i)
\end{equation}

\begin{equation}
r = \sum_{i=1}^n \alpha_i h_i
\end{equation}

where $h_i$ represents hidden states and $W_{\text{attn}}$ is a learnable parameter, and $r$ is the final representation of the input.

\paragraph{\textbf{Dual Task-Specific Heads:}} We designed parallel classification pathways for promise (Subtask~1) and evidence (Subtask~2) detection with identical architectures consisting of sequential layers:
\begin{equation}
\resizebox{0.79\hsize}{!}{
$\text{Classifier}(x) = W_2\text{GELU}(\text{LN}(\text{Dropout}(W_1x)))
$}
\end{equation}

Each classifier employs layer normalization for training stability, dropout for regularization, and GELU activation functions for improved gradient flow.

\paragraph{\textbf{Context-Enriched Representation:}} We incorporated document metadata directly into text representations by prepending structural markers:
\begin{equation}
\resizebox{0.9\hsize}{!}{
$x_{\text{enriched}} = \text{"[PAGE } p \text{] [ESG REPORT] "} + x_{\text{raw}}
$}
\end{equation}

Here, [PAGE \textit{p}] is dynamically set based on the page number of the input document, and the document-type tag ([ESG REPORT]) can vary depending on the report source, allowing the model to distinguish between different document types.

\paragraph{\textbf{Multi-objective Weighting:}} The combined training objective weights the promise and evidence subtasks differently to prioritize the more foundational promise detection task:

\begin{equation}
\mathcal{L} = 0.6 \cdot \mathcal{L}_{\text{promise}} + 0.4 \cdot \mathcal{L}_{\text{evidence}}
\end{equation}

\paragraph{{\textbf{Test-Time Augmentation:}}} For prediction, we implemented multiple forward passes with different text augmentations, averaged the probabilities across ensemble predictions, and calibrated thresholds for final binary decisions.

\section{Experimental Setup}

\subsection{Models~1 \&~2: Cross‑Validation and Feature‑Enhanced Training}
For Models~1 and~2, we implemented a~4-fold stratified cross-validation approach for data splitting during hyperparameter tuning for each subtask. The English dataset was divided using the StratifiedKFold class from the scikit-learn library\footnote{\url{https://scikit-learn.org/stable/modules/generated/sklearn.model_selection.StratifiedKFold.html}}, maintaining class distribution across folds to address class imbalance. For each trial, the data was partitioned with 75\% used for training and~25\% for validation. Validation loss was the sole optimization metric, averaged across all~4 folds for each of the 7 trials per subtask. We maintained consistent random seeds throughout all experiments to ensure reproducibility. 

For Model~2, our preprocessing approach varied by subtask, with each designed to extract task-specific linguistic features. For promise identification, we used sentiment analysis through the \texttt{Flair} package~\citep{akbik-etal-2019-flair} and promise word detection. For evidence identification, we counted concrete metrics and supporting proof terms while detecting named entities using \texttt{spaCy}\footnote{\label{spacy}\url{https://spacy.io/}}. For clarity assessment, we analyzed the prevalence of vague versus specific terminology. For the timing for verification, we extracted dates and identified timeline indicators. All enriched features were prepended to the original text as specialized tags before tokenization.

Hyperparameter optimization was performed using Optuna with the TPE sampler. We tuned the learning rate (1e-5 to 5e-5), batch size (4, 8,~12), and weight decay (0.01 to 0.3). We used early stopping with a patience of~2 epochs to prevent overfitting. For model architecture, we fine-tuned only the last two transformer layers and the classification head while freezing earlier layers. After determining optimal hyperparameters, the final model for each subtask was trained on the entire dataset.

\subsection{Model~3: Multi‑Task Learning Setup}
For Model~3, we adopted a different experimental approach to leverage the multi-task learning paradigm. We divided the English dataset using stratified sampling with a 90-10 train-validation split, ensuring a balanced representation of both promise and evidence classes. This larger training proportion was selected to provide sufficient examples for the joint learning task. The model was trained with a learning rate of~1e-5, weight decay of 0.01, and a cosine learning rate scheduler with~10\% warmup steps. To accommodate memory constraints while maintaining effective batch sizes, we implemented gradient accumulation with~16 steps and reduced sequence length to~256 tokens. Training proceeded for 5 epochs with evaluation on a held-out validation set after each epoch, with the best checkpoint saved based on the average F1 score across both tasks. For inference, we employed test-time augmentation~\cite{shanmugam2021betteraggregationtesttimeaugmentation} with~3 forward passes using random word dropout (10\%) and metadata variations, then ensemble-averaged the predictions with calibrated thresholds (0.5) for final classification.

\section{Results and Discussions}
Table~\ref{tab:model_results} presents the performance of our three models on both public and private leaderboards. The public leaderboard score is computed using approximately~33\% of the test set, while the private leaderboard score determines the final standings based on the remaining 67\%. 

Since the Combined Model only worked on Tasks~1 and~2, and Kaggle required all four subtasks for evaluation, we incorporated Task~1 and~2 predictions from our Combined Model while using our Feature-Enhanced Model for Tasks~3 and~4. 

The private scores show improvement across our models. Starting with the Base Model (0.4994), we achieved better results with the Feature-Enhanced Model (0.5094), and our Combined Subtask Model reached 0.5268, surpassing the Kaggle Baseline (0.5227). While the improvements are modest, they suggest our architectural changes and feature engineering methods are effective for promise verification tasks.

\begin{table}[ht]
    \centering
    \small
    \begin{tabular}{lcc}
        \toprule
        \textbf{Model} & \textbf{Public Score} & \textbf{Private Score} \\
        \midrule
        Kaggle Baseline & 0.5523 & 0.5227\\
        Base Model & 0.5082 & 0.4994 \\
        Feature-Enhanced Model & 0.5137 & 0.5094 \\
        Combined Subtask Model & 0.5255 & 0.5268 \\
        \bottomrule
    \end{tabular}
    \caption{Performance of our models on the SemEval-2025 Task 6 leaderboard compared to the Kaggle baseline. The public score is calculated using~33\% of the test set, while the private score reflects the final evaluation based on 67\% of the test set.}
    \label{tab:model_results}
\end{table}

Our Base Model and the Feature-Enhanced Model show a slight improvement of approximately 0.010 on the private leaderboard (from 0.4994 to 0.5094). Our minimal improvements in performance likely stem from ESG-BERT already implicitly capturing many of these patterns during domain-specific pre-training, creating a redundancy effect. Additionally, our prepending approach may have created a structural disconnect between features and relevant text spans, while the limited training data (400 instances) constrained the model's ability to learn optimal weightings for the introduced features.

The Combined Subtask Model yields the largest gain, achieving 0.5268, a~1.74\% absolute improvement over the baseline. We attribute this improvement to three factors: (1) multitask learning benefits from shared representations between promise and evidence detection, (2) attention pooling allows the model to focus on semantically relevant tokens dynamically, and (3) test-time augmentation reduces variance in prediction by ensembling multiple augmentations. However, despite achieving our highest score, the Combined Subtask Model showed only modest gains relative to its substantially increased architectural complexity and computational requirements. The limited size of our training dataset (400 instances) may have prevented the model from fully leveraging its advanced components like attention pooling and multi-objective learning, while the potential negative transfer between promise and evidence tasks may have constrained performance gains for instances where these classifications require contradictory feature attention.

\section{Conclusion}
Our work explored three model architectures for promise verification in ESG reports: a baseline ESG-BERT model, a feature-enhanced model incorporating linguistic markers, and a combined subtask model with attention pooling. The combined model achieved the best performance (0.5268 on the private leaderboard), outperforming the Kaggle baseline. Despite the challenge of class imbalance across all four subtasks, our linguistic feature extraction approach and multi-objective learning framework demonstrated effectiveness in promise verification with limited training data. 

Future work could explore incorporating cross-lingual promise verification through multilingual transformer models, integrating more advanced linguistic pattern recognition, and incorporating all classification tasks within a single multi-objective architecture to better capture interdependencies between promise identification, evidence assessment, clarity evaluation, and verification timing. Additionally, a systematic ablation study could quantify the contribution of each backbone pre-trained language model (PLM), such as BERT and RoBERTa, and each feature strategy (e.g., linguistic features, document metadata).

\bibliography{custom}
\bibliographystyle{acl_natbib}

\end{document}